\documentclass[10pt,twocolumn,letterpaper]{article}

\usepackage[accsupp]{axessibility}
\usepackage{iccv}
\usepackage{times}
\usepackage{epsfig}
\usepackage{graphicx}
\usepackage{amsmath}
\usepackage{amssymb}
\usepackage{bm}
\usepackage{bbding}
\usepackage{xcolor}

\usepackage{multirow} 
\usepackage{multicol} 
\usepackage{arydshln}

\definecolor{grey}{rgb}{0.827,0.827,0.827}
\usepackage[breaklinks=true,bookmarks=false]{hyperref}

\iccvfinalcopy 

\def\iccvPaperID{9623} 

\ificcvfinal\fi

\begin{document}

\title{Improving 3D Object Detection with Channel-wise Transformer}

\author{{Hualian Sheng}$^{1,2}\thanks{This work was done when the author was visiting Alibaba as a research intern. The code is available at \href{https://github.com/hlsheng1/CT3D}{https://github.com/hlsheng1/CT3D}}$ \quad Sijia Cai$^{2}$ \quad Yuan Liu$^{2}$ \quad
Bing Deng$^{2}$ \\ Jianqiang Huang$^{2}$ \quad Xian-Sheng Hua$^{2}$ \quad  Min-Jian Zhao$^{1}\thanks{Corresponding author.}$\\
$^{1}$ College of Information Science and Electronic Engineering, Zhejiang University  \\
$^{2}$DAMO Academy, Alibaba Group\\
{\tt\small hlsheng@zju.edu.cn,  \{stephen.csj, alen.ly, dengbing.db\}@alibaba-inc.com}\\
{\tt\small jianqiang.jqh@gmail.com, xiansheng.hxs@alibaba-inc.com, mjzhao@zju.edu.cn}
}

\maketitle
\ificcvfinal\thispagestyle{empty}\fi

\begin{abstract}
Though 3D object detection from point clouds has achieved rapid progress in recent years, the lack of flexible and high-performance proposal refinement remains a great hurdle for existing state-of-the-art two-stage detectors. Previous works on refining 3D proposals have relied on human-designed components such as keypoints sampling, set abstraction and multi-scale feature fusion to produce powerful 3D object representations. Such methods, however, have limited ability to capture rich contextual dependencies among points. In this paper, we leverage the high-quality region proposal network and a Channel-wise Transformer architecture to constitute our two-stage 3D object detection framework (CT3D) with minimal hand-crafted design. The proposed CT3D simultaneously performs 
proposal-aware embedding and channel-wise context aggregation for the point features within each proposal. Specifically, CT3D uses proposal's keypoints for spatial contextual modelling and learns attention propagation in the encoding module, mapping the proposal to point embeddings. Next, a new channel-wise decoding module enriches the query-key interaction via channel-wise re-weighting to effectively merge multi-level contexts, which contributes to more accurate object predictions. Extensive experiments demonstrate that our CT3D method has superior performance and excellent scalability. Remarkably, CT3D achieves the AP of 81.77\% in the moderate car category on the KITTI test 3D detection benchmark, outperforms state-of-the-art 3D detectors. 
\end{abstract}


\section{Introduction}\label{intro}
3D object detection from point clouds is envisioned as an indispensable part of future Autonomous Vehicle (AV). Unlike the developed 2D detection algorithms whose success is mainly due to the regular structure of image pixels, LiDAR point clouds are usually sparse, unordered and unevenly distributed. This makes the CNN-like operations not well suited to process unstructured point clouds directly. To tackle these challenges, many approaches employ voxelization or custom discretization for point clouds. Several methods~\cite{su2015multi, li2016vehicle} project point clouds to a birds-eye view (BEV) representation and apply the standard 2D convolutions, however, it will inevitably sacrifice certain geometric details which are vital for generating accurate localization. Other methods~\cite{chen2017multi,second} rasterize point clouds into a 3D voxel grid and use regular 3D CNNs to perform computation in grid space, but this category of methods suffers from computational bottleneck associated with making the grid finer. A major breakthrough in detection task on point clouds is due to the effective deep architectures for point clouds representation such as volumetric convolution~\cite{second} and permutation invariant convolution~\cite{qi2017pointnet}.

Recently, most state-of-the-art methods for 3D object detection adopt a two-stage framework consisting of 3D region proposal generation and proposal feature refinement. Notice that the most popular region proposal network (RPN) backbone~\cite{second} has achieved over 95\% recall rate on the KITTI 3D Detection Benchmark, whereas this method only achieves 78\% Average Precision (AP). The reason for such a gap stems from the difficulty in encoding an object and extracting the robust feature from 3D proposals in cases of occlusion or long-range distance. Therefore, how to effectively model geometric relationships among points and exploit accurate position information during the proposal feature refinement stage is crucial for good performance. An important family of models is PointNet~\cite{qi2017pointnet} and its variants~\cite{qi2017pointnet++,pan20203d,shi2019pointrcnn}, which use a flexible receptive field to aggregate features by local regions and permutation-invariant network. However, these methods have the drawback of involving plenty of hand-crafted designs, such as the neighbor ball radii and the grid size. Another family of models is the voxel-based methods~\cite{second,song2016deep,zheng2020cia} which use 3D convolutional kernels to gather information from neighboring voxels. But the performance of such methods is not optimal caused by the voxel quantization and sensitive to hyper-parameters. Later studies~\cite{voxelnet,shi2020pv,deng2020voxel,he2020structure} further apply the point-voxel mixed strategy to capture multi-scale features while retaining fine-grained localization but are strongly tied to the specific RPN architectures.

In this paper, we make two major contributions. First, we propose a novel end-to-end two-stage 3D objection detection framework called CT3D. Motivated by the recent Transformer-based 2D detection method DETR~\cite{carion2020end} that uses CNN backbone to extract features and encoder-decoder Transformer to enhance the RoI region features, we design our CT3D to generate 3D bounding boxes at the first stage, then learn per-proposal representation by incorporating a novel Transformer architecture with channel-wise re-weighting mechanism in decoder. The proposed framework exhibits very strong performance in terms of accuracy and efficiency, and thus can be conveniently combined with any high-quality RPN backbones.

The second contribution is the custom Transformer that offers several benefits over the traditional point/voxel-based feature aggregation mechanism. Despite the point-wise or voxel convolutions have the ability of local and global context modelling, there still have been several limitations in increasing receptive field and parameter optimization. In addition, point-cloud based 3D object detectors also have to deal with the challenging missing/noisy detections such as occlusion and distancing patterns with a few points. Self-attention in Transformers has recently emerged as a basic building block for capturing long-range interactions thus is a natural choice in acquiring context information for enriching the faraway objects or increasing the confidence of false negatives. Inspired by this idea, we initially introduce a proposal-to-point embedding to effectively encode the RPN proposal information in the encoder module. Furthermore, we exploit a channel-wise re-weighting approach to augment the standard Transformer decoder in consideration of both global and local channel-wise features for the encoded points. The purpose is to scale the feature decoding space where we can compute attention distribution over each channel dimension of key embeddings thus can enhance the expressiveness of query-key interactions. Extensive experiments show that our proposed CT3D can outperform the state-of-the-art published methods on both the KITTI dataset and the large-scale Waymo dataset.

\begin{figure*}
	\begin{center}
		\includegraphics[width=1.0\linewidth]{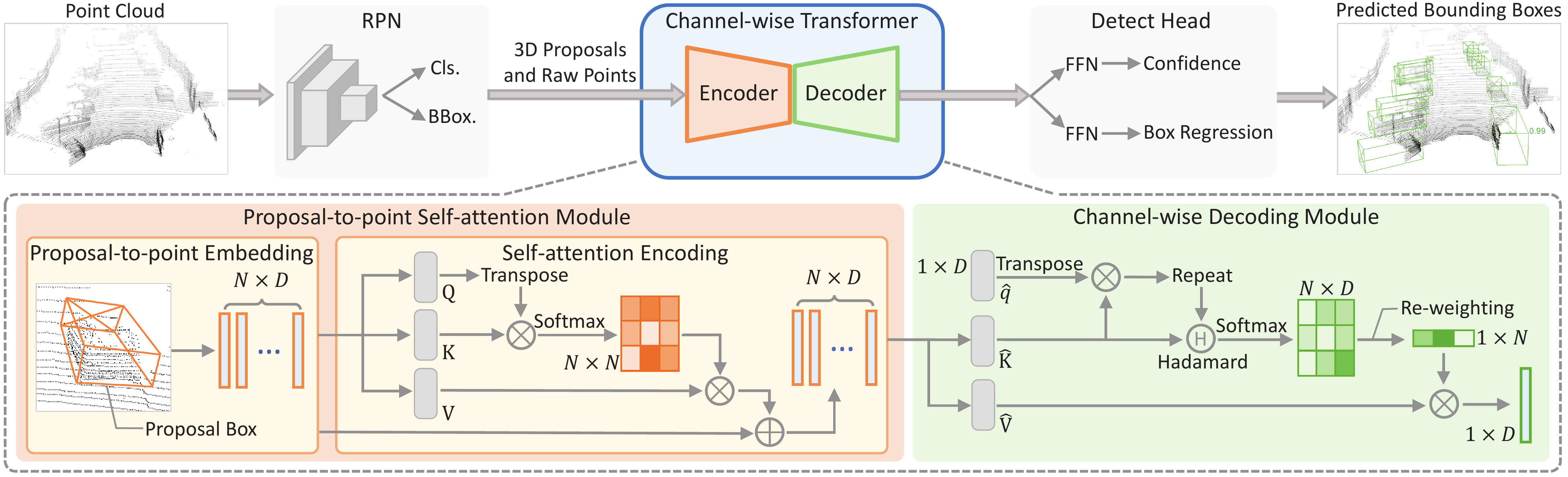}
	\end{center}
	\caption{Overview of CT3D. The raw points are first fed into the RPN for generating 3D proposals. Then the raw points along with the corresponding proposals are processed by the channel-wise Transformer composed of the proposal-to-point encoding module and the channel-wise decoding module. Specifically, the proposal-to-point encoding module is to modulate each point feature with global proposal-aware context information. After that, 
	the encoded point features  are transformed into an effective proposal feature representation by the channel-wise decoding module for confidence prediction and box regression.}
	\label{fig:TSD}
\end{figure*}

\section{Related Work}
\noindent\textbf{Point Cloud Representations for 3d Object Detection.} Recently, there has been a lot of progress on learning effective representations for the raw LiDAR point clouds. A noticeable portion of efforts are PointNet series~\cite{qi2017pointnet} which employed permutation invariant operations to aggregate the point features. F-PointNet~\cite{qi2018frustum} generated the region-level features for point clouds within each 3D frustum. PointRCNN~\cite{shi2019pointrcnn} used PointNet++~\cite{qi2017pointnet++} to segment foreground 3D points and reﬁne the proposals with the segmentation features. STD~\cite{yang2019std} further extended the proposal refinement by transferring sparse point features into dense voxel representation. Moreover, 3DSSD~\cite{yang20203dssd} improved the point-based approach with a new sampling strategy based on feature distance. However, PointNet-like architectures still present limited ability to capture local structures for LiDAR data. Another category of methods~\cite{chen2017multi,lang2019pointpillars, yang2018hdnet, yang2018pixor, su2015multi, li2016vehicle,ku2018joint,liang2019multi,liang2018deep} aimed to voxelize the unstructured point clouds as a regular 2D/3D grid over which conventional CNNs can be easily applied. Pioneer work~\cite{chen2017multi} encoded the point clouds as 2D bird-view feature maps to generate highly accurate 3D candidate boxes, motivating many efﬁcient bird-view representation-based methods. VoxelNet~\cite{voxelnet} transformed the points to form a compact feature representation. SECOND~\cite{second} introduced 3D sparse convolution for efﬁcient 3D voxel processing. These voxel-based methods are still focused on the subdivision of a volume rather than adaptively modelling local geometric structure. Furthermore, various point-voxel based methods have been proposed for multi-scale feature aggregation. SA-SSD~\cite{he2020structure} presented an auxiliary network on the basis of 3D voxel CNN. PV-RCNN~\cite{shi2020pv} and its variant VoxelRCNN~\cite{deng2020voxel} adopted 3D voxel CNN as RPN to generate high-quality proposals and then utilize PointNet to aggregate the voxel features around the grids. Nevertheless, these hybrid methods require plenty of hand-crafted feature designs.

\noindent\textbf{Transformers for object detection.} A new paradigm for object detection has recently evolved due to the success of Transformers in many computer vision fields~\cite{carion2020end,zhu2020deformable,dosovitskiy2020image,guo2020pct,engel2020point}. Since Transformer models are very effective at learning local context-aware representations, DETR~\cite{carion2020end} viewed the detection as a set prediction problem and employed Transformer with parallel decoding to detect objects in 2D image. A variant of DETR~\cite{zhu2020deformable} further developed a deformable attention module to employ cross-scale aggregation. For point clouds, recent methods~\cite{guo2020pct,engel2020point} also explored to use self-attention for classiﬁcation and segmentation tasks. 

\section{CT3D for 3D Object Detection}
Given proposals generated by the widely used RPN backbones like 3D voxel CNN~\cite{second}, current state-of-the-art proposal refinement approaches~\cite{shi2020pv,deng2020voxel} focus on refining the intermediate multi-stage voxel features extracted by the convolution layers, suffering the difficulties of extra hyper-parameter optimization and designing generalized models. We believe that the raw points with precise position information are sufficient for refining the detection proposals.
Bearing this view in mind, we construct our CT3D framework by deploying a well-designed Transformer on top of a RPN network to directly utilize the raw point clouds. Specifically, the whole CT3D detection framework is composed of three parts, \textit{i.e., }a RPN backbone for proposal generation, a channel-wise Transformer for proposal feature refinement and a detect head for object predictions. Figure~\ref{fig:TSD} illustrates an overview of our CT3D framework.

\subsection{RPN for 3D Proposal Generation}
Starting from the point clouds $\textbf{P}$ with 3-dimension coordinates and $C$-dimension point features, the predicted 3D bounding box generated by RPN consists of center coordinate $\bm{p}^c = [x^{c}, y^{c}, z^{c}]$, length $l^c$, width $w^c$, height $h^c$, and orientation $\theta^{c}$. In this paper, we adopt the 3D voxel CNN SECOND~\cite{second} as our default RPN due to its  high efficiency and accuracy. Note that any high-quality RPN should be readily replaceable in our framework and is amenable to training via an end-to-end manner.

\subsection{Proposal-to-point Encoding Module}\label{encoder}
To refine the generated RPN proposals, we adopt a two-step strategy. Specifically, the first proposal-to-point embedding step maps the proposal to point features, then the second self-attention encoding step is to refine point features via modelling the relative relationships among points within the corresponding proposal.

\noindent\textbf{Proposal-to-point Embedding.} Given the proposals generated by RPN, we delimit out a scaled RoI area in point clouds according to the proposal. This aims to compensate the deviation between the proposal and the corresponding ground-truth box by wrapping all object points as much as possible. Specifically, the scaled RoI area is a cylindrical with unlimited height and a radius $r =\alpha \sqrt{(\frac{l^c}{2})^2 + (\frac{w^c}{2})^2}$, where $\alpha$ is a hyper-parameter, and $l$, $w$ denote the length and width of the proposal, respectively. Hereinafter, the randomly sampled $N = 256$ points within the scaled RoIs ($\mathcal{N} = \{\bm{p}_1, \dots, \bm{p}_N\}$ ) are taken out for further processing. 

At first, we calculate the relative coordinates between each sampled point and the center point of the proposal for unifying the input distance feature, denoted as $\Delta \bm{p}_i^c = \bm{p}_i - \bm{p}^c, \forall \bm{p}_i \in \mathcal{N}$. A straightforward thought is to directly concatenate the proposal information into each point feature, \textit{i.e., }$[\Delta \bm{p}_i^c, l^c, w^c, h^c, \theta^c, f_i^r]$, where $f_i^r$ is the raw point feature such as reflection. However, the size-orientation representation for proposal yields only modest performance as the Transformer encoder might be less effective to reorient in accord with above-mentioned geometric information.

\begin{figure}[t]
	\begin{center}
		\includegraphics[width=1.0\linewidth]{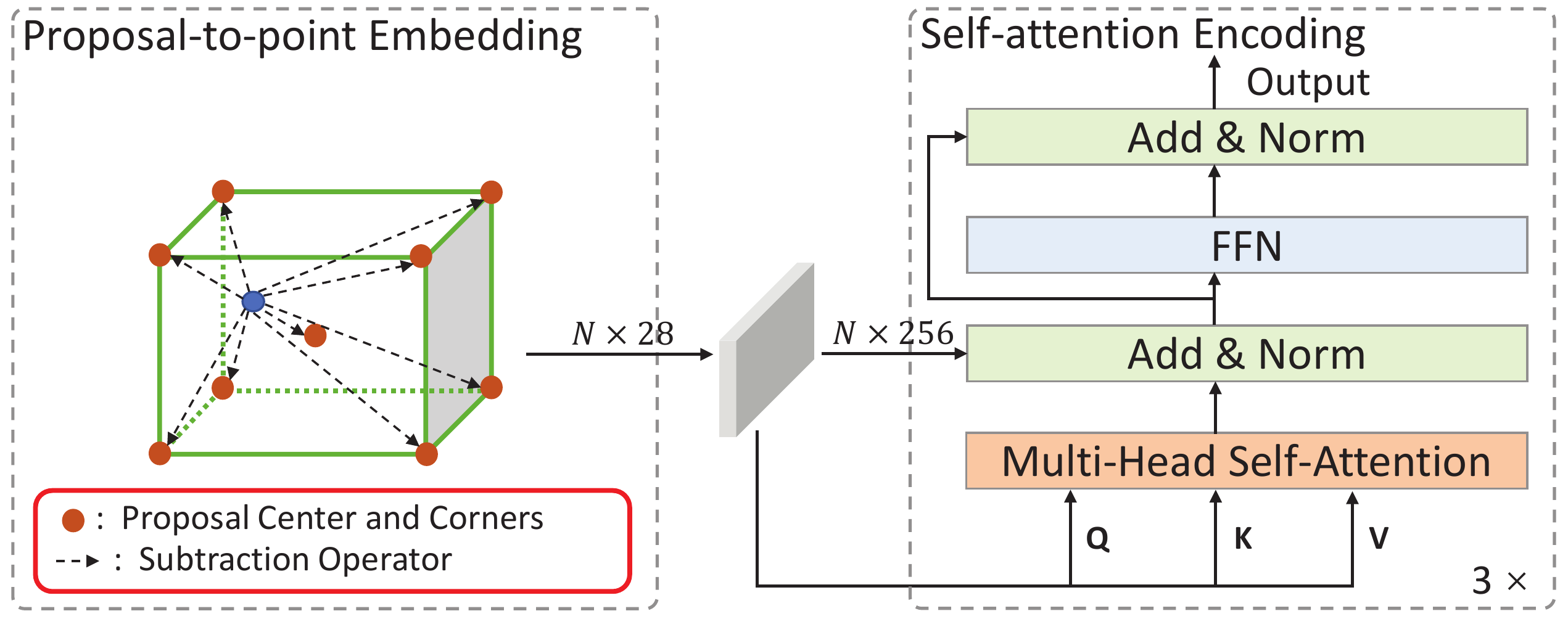}
	\end{center}
	\caption{Proposal-to-point encoding. The location features of raw point clouds are first modulated by the proposal information (center and corners) via subtraction operator. Then, the resulting point features are refined by the proposal-aware encoding module with multi-head self-attention mechanism.}
	\label{fig:encoder}
\end{figure}

It is noteworthy that the keypoints usually offer more explicit geometry property in detection tasks~\cite{zhou2019bottom, law2018cornernet}, we propose a novel keypoints subtraction strategy to compute the relative coordinates between each point and the eight corner points of the corresponding proposal. The calculated relative coordinates are  $\Delta \bm{p}_i^j = \bm{p}_i - \bm{p}^j, j = 1,\dots, 8$, where $ \bm{p}^j$ is the coordinate of the $j$-th corner point. 
Note that $l^c, w^c, h^c$ and $\theta^c$ disappear but are contained in different dimensions of distance information. Through this way, the newly generated relative coordinates $\Delta \bm{p}_i^j$ can be viewed as a better representation of proposal information.
As shown in the left part of Figure~\ref{fig:encoder}, for each point $\bm{p}_i$, the proposal-guided point feature can be expressed as:
\begin{align}
	\bm{f}_i = \mathcal{A}([\Delta \bm{p}_i^c, \Delta \bm{p}_i^1, \dots, \Delta \bm{p}_i^8, f_i^r]) \in \mathbb{R}^D,
\end{align}
where $ \mathcal{A}(\cdot)$ is a linear projection layer to map point feature into a high-dimensional embedding.

\noindent\textbf{Self-attention Encoding.} The embeded point features are then fed into the multi-head self-attention layer, followed by a feed-forward network (FFN) with residual structure, to encode rich contextual relationships and point dependencies in proposal for refining point features. As shown in the right part of Figure~\ref{fig:encoder}, this self-attention encoding scheme shares almost the same structure as the original NLP Transformer encoder, except for the position embedding since it is already included in the point features. Reader can refer to~\cite{vaswani2017attention} for more details. Denote $\mathbf{X} = [\bm{f}_1^T,\dots,\bm{f}_N^T]^T \in \mathbb{R}^{N \times D}$ as the embedded point features with the dimension $D$, we have $\mathbf{Q} = \mathbf{W}_q\mathbf{X};\ \  \mathbf{K} =  \mathbf{W}_k\mathbf{X};\ \  \mathbf{V} = \mathbf{W}_v\mathbf{X},$ where $\mathbf{W}_q, \mathbf{W}_k, \mathbf{W}_v \in \mathbb{R}^{N \times N}$ are linear projections, and $\mathbf{Q}, \mathbf{K}$ and $\mathbf{V}$ are so-called query, key and value embeddings. These three embeddings are then processed by multi-head self-attention mechanism. In a $H$-head attention situation, $\mathbf{Q}, \mathbf{K}$ and $\mathbf{V}$ are further divided into $\mathbf{Q}=[\mathbf{Q}_1, \dots, \mathbf{Q}_H]$, $\mathbf{K}=[\mathbf{K}_1, \dots,\mathbf{K}_H]$, and $\mathbf{V}=[\mathbf{V}_1, \dots, \mathbf{V}_H]$, where $\mathbf{Q}_h, \mathbf{K}_h, \mathbf{V}_h  \in \mathbb{R}^{N \times D'}, \forall h=1,\dots,H$, and $D' = \frac{D}{H}$. The output after multi-head self-attention is given by:
\begin{align}\label{equ:self-att}
	\text{S}^{\text{(att)}}(\mathbf{Q} ,\mathbf{K} ,\mathbf{V} ) = \bigg[\sigma\big(\frac{\mathbf{Q}_h\mathbf{K}_h^T}{\sqrt{D'}}\big)\cdot\mathbf{V}_h\bigg],  h = 1,\dots,H,
\end{align}
where $\sigma(\cdot)$ is \textit{softmax} function. Hereinafter, applying a simple FFN and residual operator, the result is as follows:
\begin{align}
	\text{S}^{\text{(emb)}}(\mathbf{X}) = \mathcal{Z}(\mathcal{F}(\mathcal{Z}(\text{S}^{\text{(att)}}(\mathbf{Q} ,\mathbf{K} ,\mathbf{V} )))),
\end{align}
where $\mathcal{Z}(\cdot)$ denotes add and normalization operator, $\mathcal{F}(\cdot)$ denotes a FFN with two linear layers and one \textit{Relu} activation. We observe that a stack of 3 identical self-attention encoding modules is ideal for our CT3D framework.

\subsection{Channel-wise Decoding Module}\label{decoder}
In this subsection, we manage to decode all point features (\textit{i.e.,} $\hat{\mathbf{X}}$) from the encoder module into a global representation, which is further processed by FFNs for the final detection predictions. Different from the standard Transformer decoder, which transforms $M$ multiple query embeddings using self- and encoder-decoder attention mechanism, our decoder only manipulates one query embedding according to the following two facts:
\begin{itemize}
	\setlength{\itemindent}{0em}
	\item  $M$ query embeddings suffer high memory latency, especially for processing with numbers of proposals.
	\item $M$ query embeddings are usually independently transformed into $M$ words or objects, while our proposal refinement model only needs one prediction.
\end{itemize}

Generally, the final proposal representation after decoder can be regarded as a weighted sum of all point features, our key motivation is to determine the decoding weights that are dedicated for each point. In below, we first analyze the standard decoding scheme, and then develop an improved decoding scheme to acquire more effective decoding weights.

\begin{figure}[t]
	\begin{center}
		\includegraphics[width=1.0\linewidth]{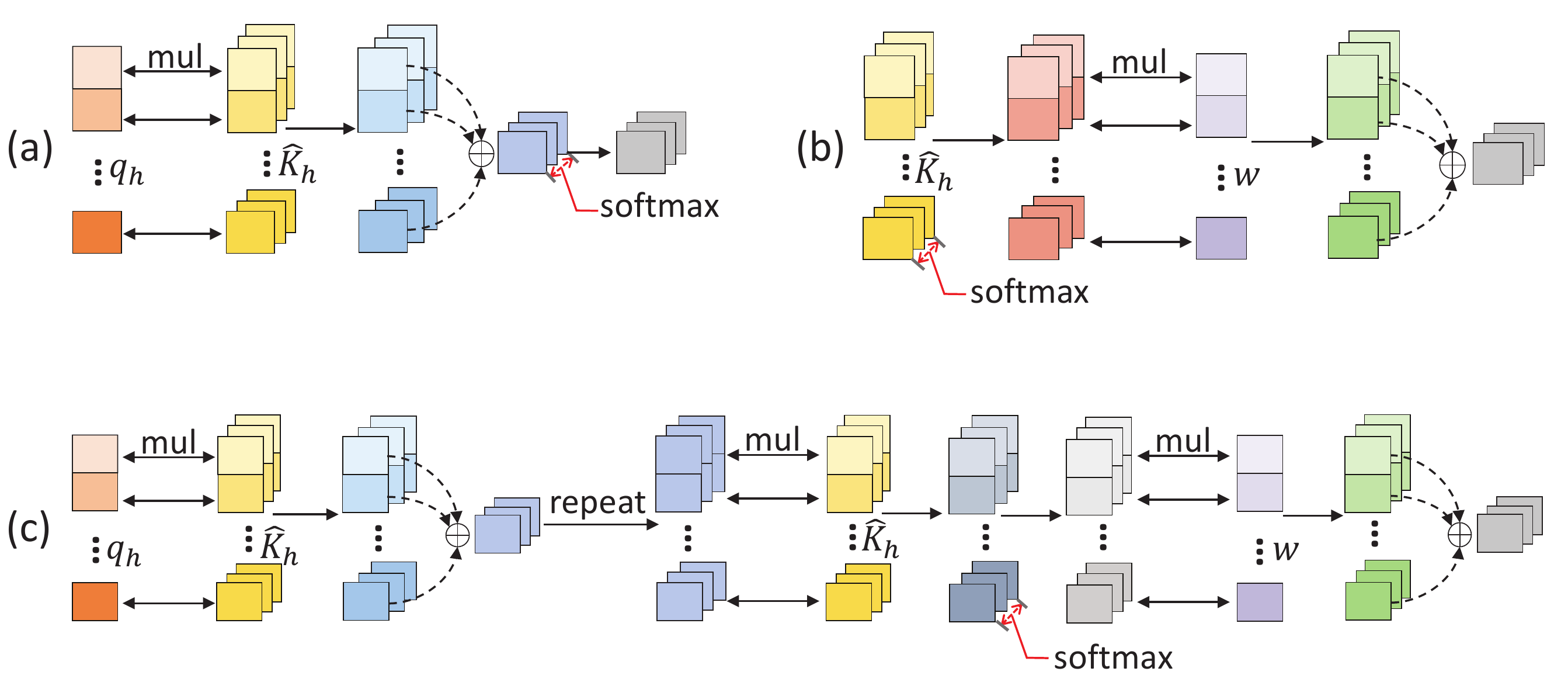}
	\end{center}
	\caption{Illustration of the different decoding schemes: (a) Standard decoding; (b) Channel-wise re-weighting; (c) Extended channel-wise re-weighting.}
	\label{fig:channel_wise}
\end{figure}

\noindent\textbf{Standard Decoding.} The standard decoding scheme utilizes a learnable vector (\textit{i.e., }query embedding) of dimension $D$ to aggregate the point features across all channels. As shown in Figure~\ref{fig:channel_wise}(a), the final decoding weight vector for all point features in each attention head is:
\begin{align}
    \bm{w}_h^{(S)} =\sigma\big(\frac{\hat{\bm{q}}_h\hat{\mathbf{K}}_h^T}{\sqrt{D'}}\big), h = 1, \dots, H,
\end{align}
where $\hat{\mathbf{K}_{h}}$ is the key embeddings of $h$-th head computed by the projection of encoder output, and $\hat{\bm{q}}_h$ is the corresponding query embedding. Note that each value of vector $\hat{\bm{q}}_h\hat{\mathbf{K}}_h^T$ can be viewed as the global aggregation for individual point (\textit{i.e.}, each key embedding), and the subsequent \textit{softmax} function assigns the decoding value for each point according to the probability in the normalized vector. Consequently, the values in decoding weight vector are derived from simple global aggregation and lack the local channel-wise modelling, which is essential to learn 3D surface structures of point clouds because different channels usually exhibit strong geometric relationships in point clouds.

\noindent\textbf{Channel-wise Re-weighting.} In order to emphasize the channel-wise information for key embeddings $\hat{\mathbf{K}}_h^T$, a straightforward solution is to compute the decoding weight vector for points based on all the channels of $\hat{\mathbf{K}}_h^T$. That is, we generate $D$ different decoding weight vectors for each channel to obtain $D$ decoding values. Further, a linear projection is introduced for these $D$ decoding values to form a united channel-wise decoding vector. As shown in Figure~\ref{fig:channel_wise}(b), this new channel-wise re-weighting for decoding weight vector can be summarized as:
\begin{align}
    \bm{w}_h^{(C)} =\bm{s} \cdot \hat{\sigma}\big(\frac{\hat{\mathbf{K}}_h^T}{\sqrt{D'}}\big), h = 1, \dots, H,
\end{align}
where $\bm{s}$ is a linear projection that compresses $D'$ number of decoding values into a re-weighting scalar, $\hat{\sigma}(\cdot)$ computes the \textit{softmax} along the $N$ dimension.
However, the decoding weights computed by $\hat{\sigma}(\cdot)$ are associated with each channel, and thus ignore the global aggregation of each point. Therefore, we can conclude that the standard decoding scheme focuses on global aggregation while the channel-wise re-weighting scheme concentrates on the channel-wise local aggregation. To combine their characteristics, we propose an extended channel-wise re-weighting scheme as below.

\noindent\textbf{Extended Channel-wise Re-weighting.} Specifically, we first repeat the matrix product of query embedding and key embeddings to spread the spatial information into each channel, and the output is then multiplied element-wise with the key embeddings for keeping the channel differences. As illustrated in Figure~\ref{fig:channel_wise} (c), this novel extended channel-wise re-weighting scheme generates the following decoding weight vector for all the points:
\begin{align}
	\bm{w}_h^{(EC)} = \bm{s}\cdot\hat{\sigma}\big(\frac{\rho(\hat{\bm{q}}_h\hat{\mathbf{K}}_h^T)\odot \hat{\mathbf{K}}_h^T}{\sqrt{D'}}\big), h = 1, \dots, H,
\end{align}
where $\rho(\cdot)$ is a repeat operator makes $\mathbb{R}^{1\times N} \to \mathbb{R}^{D'\times N}$. In this way, we can not only maintain the global information as compared to the channel-wise re-weighting scheme, but also enrich the local and detailed channel interactions as compared to the standard decoding scheme. Besides, this extended channel-wise re-weighting only brings 1K+ (Bytes) increase as compared to the other two schemes. As a result, the final decoded proposal representation can be described as follows:
\begin{align}
    \bm{y} = [\bm{w}_1^{(EC)}\cdot\hat{\mathbf{V}}_1, \dots, \bm{w}_H^{(EC)}\cdot\hat{\mathbf{V}}_H],
\end{align}
where the value embeddings $\hat{\mathbf{V}}$ is the linear projection obtained from $\hat{\mathbf{X}}$.

\subsection{Detect head and Training Targets}
In the previous steps, the input point features are summarized into a $D$-dimension vector $\bm{y}$, which is then fed into two FFNs for predicting the confidence and the box residuals relative to the input 3D proposal, respectively.

To output the confidence, training targets are set as the 3D IoU between the 3D proposals and their corresponding ground-truth boxes. Given the IoU of the 3D proposal and its corresponding ground-truth box, we follow~\cite{jiang2018acquisition, shi2019pointrcnn,shi2020pv} to assign the confidence prediction target, which is shown as:
\begin{align}
	 c^t = \min\bigg(1, \max \big(0, \frac{\text{IoU} - \alpha_B}{\alpha_F - \alpha_B}\big)\bigg) ,
\end{align}
where $\alpha_F$ and $\alpha_B$ are the foreground and background IoU thresholds, respectively. Besides, regression targets (superscript $t$) are encoded by proposals and their corresponding ground-truth boxes (superscript $g$), given by:
\begin{align}
	\notag&x^t = \frac{x^g - x^c}{d}, y^t = \frac{y^g - y^c}{d}, z^t = \frac{z^g - z^c}{h^c}, \\
	\notag &l^t = \log{(\frac{l^g}{l^c})}, w^t = \log{(\frac{w^g}{w^c})}, h^t = \log{(\frac{h^g}{h^c})}, \\
	&\theta^t = \theta^g - \theta^c,
\end{align}
where $d = \sqrt{(l^c)^2 + (w^c)^2}$ is the diagonal of the base of the proposal box.

\subsection{Training Losses}
We adopt an end-to-end strategy to train  CT3D. Hence, the overall training loss is the summation of the RPN loss, the confidence prediction loss, and the box regression loss, which is presented:
\begin{align}
	\mathcal{L} = \mathcal{L}_{\text{RPN}} + \mathcal{L}_{\text{conf}} + \mathcal{L}_{\text{reg}}.
\end{align}

Here, the binary cross entropy loss~\cite{jiang2018acquisition, yang2018pixor} is exploited for the predicted confidence $c$ to compute the IoU-guided confidence loss:
\begin{align}
	\mathcal{L}_{\text{conf}} = -c^t\log{(c)} - (1-c^t)\log{(1-c)}.
\end{align}

Moreover, the box regression loss~\cite{yang2018pixor, second} adopts:
\begin{align}
	\mathcal{L}_{\text{reg}} =\mathbb{I}(\text{IoU} \geq \alpha_R)\sum_{\mu \in {x, y, z, l, w, h, \theta}}{\mathcal{L}_{\text{smooth-L1}}}(\mu, \mu^t),
\end{align}
where $\mathbb{I}(\text{IoU} \geq \alpha_R)$ indicates that only proposals with $\text{IoU} \geq \alpha_R$ contribute to the regression loss.

\section{Experiments}
In this section, we evaluate our CT3D on two public datasets, KITTI~\cite{geiger2013vision} and Waymo~\cite{ngiam2019starnet,zhou2020end}. Furthermore, we conduct comprehensive ablation studies to verify the effectiveness of each module in CT3D.

\subsection{Dataset}
\noindent\textbf{KITTI Dataset.} 
KITTI dataset officially contains 7,481 training LiDAR samples and 7,518 testing LiDAR samples. Following the previous work~\cite{chen20153d}, we split the original training data into 3,712 training samples and 3,769 validation samples for experimental studies.

\noindent\textbf{Waymo Dataset.} 
Waymo dataset consists of 798 training sequences with around 158,361 LiDAR samples, and 202 validation sequences with 40,077 LiDAR samples. This large-scale Waymo dataset detection task is more challenging due to its various autonomous driving scenarios~\cite{zhou2020end}.

\subsection{Implementation Details}
\noindent\textbf{RPN.}
We adopt SECOND~\cite{second} as our RPN due to its high-quality proposals and fast speed of inference. For the KITTI dataset, the $X, Y, Z$ axis ranges are set as $(0, 70.4)$, $(-40, 40)$, $(-3, 1)$, and the voxel size is set as $(0.05\text{m}, 0.05\text{m}, 0.1\text{m})$ in $(X\text{-axis}, Y\text{-axis}, Z\text{-axis})$. For the Waymo dataset, the corresponding axis ranges are $(-75.2, 75.2)$, $(-75.2, 75.2)$, $(-2, 4)$, and the voxel size is $(0.1\text{m}, 0.1\text{m}, 0.15\text{m})$. $\mathcal{L}_{\text{RPN}}$ consists of the Focal-Loss classification branch and the Smooth-L1-Loss based regression branch. Please
refer to OpenPCDet~\cite{openpcdet2020} for more details since we conduct our experiments with this toolbox.

\begin{table}[t]  
	\footnotesize
	\setlength\tabcolsep{5pt}
	\begin{center}
		\begin{tabular}{c|c||ccc}
			\hline
			\multirow{2}*{Method}  & \multirow{1}*{Par.}  & \multicolumn{3}{c}{3D Detection - Car}\\ 
			\cline{3-5}
			& (M) & Easy & Mod. & Hard \\
			\hline 
			\multicolumn{5}{c}{\textit{LiDAR \& RGB}}\\
			\hline
			MV3D, \textit{CVPR 2017}~\cite{chen2017multi} & - & 74.97 & 63.63 & 54.00\\
			ContFuse, \textit{ECCV 2018}~\cite{liang2018deep} & -  & 83.68 & 68.78 & 61.67 \\
			AVOD-FPN, \textit{IROS 2018}~\cite{ku2018joint} & -  &83.07 & 71.76 & 65.73 \\
			F-PointNet, \textit{CVPR 2018}~\cite{qi2018frustum}  &  40 & 82.19 & 69.79 & 60.59 \\
			UberATG-MMF, \textit{CVPR 2019}~\cite{liang2019multi} &-&88.40& 77.43 &70.22\\
			3D-CVF at SPA, \textit{ECCV 2020}~\cite{yoo20203d} &-&89.20 & 80.05 & 73.11\\
			CLOCs, \textit{IROS 2020}~\cite{pang2020clocs} & - & 88.94 & 80.67 & 77.15\\
			\hline
			\multicolumn{5}{c}{\textit{LiDAR only}}\\
			\hline
			SECOND, \textit{Sensor 2018}~\cite{second} & 20 & 83.34 & 72.55 & 65.82\\
			PointPillars, \textit{CVPR 2019}~\cite{lang2019pointpillars} & 18&  82.58 & 74.31 & 68.99\\
			STD, \textit{ICCV 2019}~\cite{yang2019std} & - & 87.95 & 79.71 & 75.09\\
			PointRCNN, \textit{CVPR 2019}~\cite{shi2019pointrcnn} & 16 & 86.96 & 75.64 & 70.70\\
			3D IoU Loss, \textit{3DV 2019}~\cite{zhou2019iou} &- & 86.16 & 76.50 & 71.39\\
			Part-$A^2$, \textit{PAMI 2020}~\cite{shi2020points} &226 & 87.81 & 78.49 & 73.51\\
			SA-SSD, \textit{CVPR 2020}~\cite{he2020structure} & 40.8 & 88.75 & 79.79 & 74.16\\
			3DSSD, \textit{CVPR 2020}~\cite{yang20203dssd} &-&88.36 &79.57& 74.55\\
			PV-RCNN, \textit{CVPR 2020}~\cite{shi2020pv} & 50 & 90.25 & 81.43 & 76.82\\
			Voxel-RCNN, \textit{AAAI 2021}~\cite{deng2020voxel} & 28 & \textbf{90.90} & 81.62  &	77.06\\ 
			\hline
			\hline
			CT3D (Ours) & 30 & 87.83 & \textbf{81.77} & \textbf{77.16} \\
			\hline
		\end{tabular}
	\end{center}
	\caption{Performance comparisons with state-of-the-art methods on the KITTI \textit{test} set. All results are reported by the average precision with 0.7 IoU threshold and 40 recall positions.}
	\label{table:kitti_test}
\end{table}

\begin{table}[t]  
	\footnotesize
	\begin{center}
		\begin{tabular}{c||ccc}
			\hline
			\multirow{2}*{Method}   & \multicolumn{3}{c}{3D Detection - Car}\\ 
			\cline{2-4}
			&  Easy & Mod. & Hard \\
			\hline 
			\multicolumn{4}{c}{\textit{LiDAR \& RGB}}\\
			\hline
			MV3D, \textit{CVPR 2017}~\cite{chen2017multi} & 71.29 & 62.68 & 56.56\\
			ContFuse, \textit{ECCV 2018}~\cite{liang2018deep}  & - &73.25 & -\\
			AVOD-FPN, \textit{IROS 2018}~\cite{ku2018joint}  & - & 74.44 & -\\
			F-PointNet, \textit{CVPR 2018}~\cite{qi2018frustum}   &  83.76 & 70.92 &63.65\\
			3D-CVF at SPA, \textit{ECCV 2020}~\cite{yoo20203d} & 89.67 & 79.88 & 78.47\\
			\hline
			\multicolumn{4}{c}{\textit{LiDAR only}}\\
			\hline
			SECOND, \textit{Sensor 2018}~\cite{second}  & 88.61 & 78.62 & 77.22\\
			PointPillars, \textit{CVPR 2019}~\cite{lang2019pointpillars}  &  86.62 & 76.06 & 68.91\\
			STD, \textit{ICCV 2019}~\cite{yang2019std}& 89.70 & 79.80 & \textbf{79.30}\\
			PointRCNN, \textit{CVPR 2019}~\cite{shi2019pointrcnn} & 88.88 & 78.63 & 77.38\\
			SA-SSD, \textit{CVPR 2020}~\cite{he2020structure} & \textbf{90.15} & 79.91 & 78.78\\
			3DSSD, \textit{CVPR 2020}~\cite{yang20203dssd} &89.71 &79.45 &78.67\\
			PV-RCNN, \textit{CVPR 2020}~\cite{shi2020pv} & 89.35 & 83.69 & 78.70 \\
			Voxel-RCNN, \textit{AAAI 2021}~\cite{deng2020voxel}  & 89.41 & 84.52 & 78.93\\
			\hline
			\hline
			CT3D (Ours) & 89.54 & \textbf{86.06} & 78.99 \\
			\hline
		\end{tabular}
	\end{center}
	\caption{Performance comparisons with state-of-the-art methods on the KITTI \textit{val} set. All results are reported by the average precision with 0.7 IoU threshold and 11 recall positions.}
	\label{table:kitti_val}
\end{table}

\begin{table}[t]  
	\footnotesize
	\begin{center}
		\begin{tabular}{c||ccc|ccc}
			\hline
			\multicolumn{1}{c||}{IoU}  & \multicolumn{3}{c|}{BEV Detection}  & \multicolumn{3}{c}{3D Detection}\\ 
			\cline{2-7}
			\multicolumn{1}{c||}{Thr.} & Easy & Mod. & Hard & Easy & Mod. & Hard\\
			\hline
			0.7 & 96.14 & 91.88 & 89.63 & 92.85 & 85.82 & 83.46\\
			\hline
		\end{tabular}
	\end{center}
	\caption{Performance of our CT3D on the KITTI \textit{val} set with AP calculated by 40 recall positions for \textit{car} category. }
	\label{table:add_bev_val}
\end{table}

\begin{table}[t]  
	\footnotesize
	\begin{center}
	\begin{tabular}{c||ccc|ccc}
		\hline
		\multicolumn{1}{c||}{IoU}  & \multicolumn{3}{c|}{Pedestrian}  & \multicolumn{3}{c}{Cyclist}\\ 
		\cline{2-7}
		\multicolumn{1}{c||}{Thr.} & Easy & Mod. & Hard & Easy & Mod. & Hard\\
		\hline
		0.5 & 65.73 & 58.56 & 53.04 & 91.99 & 71.60 & 67.34\\
		\hline
	\end{tabular}
\end{center}
	\caption{Performance for \textit{pedestrian} and \textit{cyclist} on the KITTI. }
	\label{table:ped_cyc}
\end{table}

\begin{table*}[tp]  
	\footnotesize
	\begin{center}
		\begin{tabular}{c|c||cccc|cccc}
			\hline
			\multirow{2}*{Difficulty}  & \multirow{2}*{Method}  & \multicolumn{4}{c|}{3D Detection - Vehicle} & \multicolumn{4}{c}{BEV Detection - Vehicle}\\ 
			\cline{3-10}
			& & Overall & 0-30m & 30-50m & 50m-Inf & Overall & 0-30m & 30-50m & 50m-Inf\\
			\hline
			\multirow{6}*{LEVEL\_1} & PointPillar, \textit{CVPR 2019}~\cite{lang2019pointpillars} & 56.62 & 81.01 & 51.75 & 27.94 & 75.57 & 92.10 & 74.06 & 55.47\\
			& MVF, \textit{CoRL 2020}~\cite{zhou2020end} & 62.93 & 86.30 & 60.02 & 36.02 & 80.40 & 93.59 & 79.21 & 63.09\\
			& Pillar-OD, \textit{arXiv 2020}~\cite{wang2020pillar} & 69.80 & 88.53 & 66.50 & 42.93 & 87.11 & 95.78 & 84.87 & 72.12\\
			& PV-RCNN, \textit{CVPR 2020}~\cite{shi2020pv} & 70.30 & 91.92 & 69.21 & 42.17 & 82.96 & 97.35 & 82.99 & 64.97\\
			& Voxel-RCNN, \textit{AAAI 2021}~\cite{deng2020voxel} & 75.59 & 92.49 & 74.09 & 53.15 & 88.19 & 97.62 & 87.34 & 77.70\\
			& CT3D (Ours) & \textbf{76.30} & \textbf{92.51} & \textbf{75.07} & \textbf{55.36} & \textbf{90.50} & \textbf{97.64} & \textbf{88.06} & \textbf{78.89}\\
			\hline
			\multirow{3}*{LEVEL\_2} & PV-RCNN, \textit{CVPR 2020}~\cite{shi2020pv}  & 65.36 & 91.58 & 65.13 & 36.46 & 77.45 & 94.64 & 80.39 & 55.39\\
			& Voxel-RCNN, \textit{AAAI 2021}~\cite{deng2020voxel} & 66.59 & 91.74 & 67.89 & 40.80 & 81.07 & 96.99 & 81.37 & 63.26\\
			& CT3D (Ours) & \textbf{69.04} & \textbf{91.76} & \textbf{68.93} & \textbf{42.60} & \textbf{81.74} & \textbf{97.05} & \textbf{82.22} & \textbf{64.34}\\
			\hline
		\end{tabular}
	\end{center}
	\caption{Performance comparisons with state-of-the-art methods on the Waymo dataset with 202 validation sequences ($\sim$ 40k samples) for vehicle detection.}
	\label{table:waymo}
\end{table*}

\noindent\textbf{Training Details.}
We use 8 $\text{V}100$ GPUs to train the entire network with batch size $24$ for the KITTI dataset and batch size $16$ for Waymo dataset. For the encoder and decoder modules of channel-wise transformer, we set $\alpha = 1.2$ and $H = 4$. For training targets, we set $\alpha_F = 0.75, \alpha_B = 0.25, \alpha_R = 0.55$, respectively.  The whole CT3D framework is trained end-to-end from scratch with ADAM optimizer for 100 epochs. We adopt cosine annealing learning rate strategy for our learning rate decay, and the maximum of leaning rate is 0.001. In the training stage, only 128 proposals are randomly selected to calculate the confidence loss while 64 ($\text{IoU} \geq \alpha_R$) proposals are selected to calculate the regression loss. In the inference stage, top-100 proposals are selected for the final prediction.

\subsection{Detection Results on the KITTI Dataset}
We compare our CT3D  with state-of-the-art methods on both the KITTI \textit{test} and \textit{val} sets with 0.7 IoU threshold. For our test submission, all the released training data is used to train the model. Following ~\cite{shi2019pointrcnn,shi2020pv,deng2020voxel,he2020structure}, the average precision (AP) for \textit{test} set is calculated with 40 recall positions, while the AP for \textit{val} set is calculated with 11 recall positions when compared to the previous methods\footnote[1]{The setting of  AP calculation is modified from 11 recall positions to 40 recall positions on 08.10.2019. For fair comparison with previous methods, we exploit the 11 recall setting on \textit{val} set.}.

\begin{figure*}
	\begin{center}
		\includegraphics[width=1.0\linewidth]{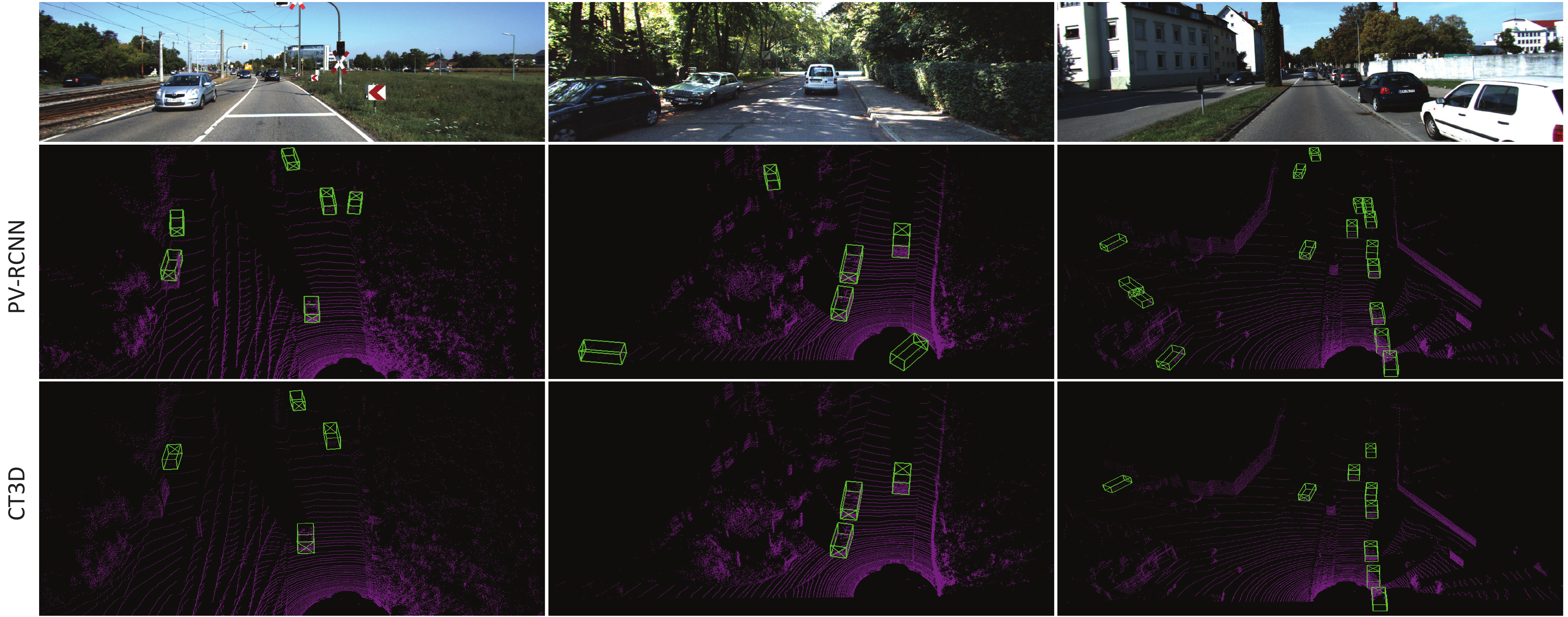}
	\end{center}
	\caption{Qualitative comparison results of 3D object detection on the KITTI \textit{test} set. Our CT3D enables more reasonable and accurate detection as compared to the PV-RCNN.}
	\label{fig:cmp}
\end{figure*}

\noindent\textbf{Performance Comparisons.}
Table~\ref{table:kitti_test} illustrates the performance comparisons between our method and state-of-the-art methods on the official KITTI \textit{test} server. It shows  CT3D achieves the best performance on \textit{moderate} and \textit{hard} levels for \textit{car} detection on both \textit{LiDAR only} and \textit{Lidar\&RGB} modalities, especially for the most important \textit{moderate} level~\cite{geiger2012we}. Compared with the newest released PV-RCNN which shares the same RPN (\textit{i.e., }SECOND) as ours, CT3D achieves better performance while requiring about 1/3 times of parameters for refinement. Besides, as shown in Figure~\ref{fig:cmp}, CT3D presents much better visualization performance as compared to the PV-RCNN. This significant improvement mainly comes from the fact that CT3D processes the raw points in refinement stage rather than relying on  human-specified designs and sub-optimal intermediate features. Note the AP on \textit{easy} level of our CT3D is comparatively worse, there might be two reasons. First, we only sample 256 raw point within each proposal for all levels even the proposals in \textit{easy} level usually have a much larger number of points. Second, we observe that KITTI exhibits large distribution differences between \textit{trainval} and \textit{test} sets.

For further  validation, we conduct comparisons with previous methods on the KITTI \textit{val} set.  It shows that our CT3D outperforms all the other methods with a large margin, leading the state-of-the-art method Voxel-RCNN by 1.54\% on \textit{moderate} level, and achieves the competitive result on \textit{easy} level. This improvement also verifies the effectiveness of our method, indicating our CT3D could better model the context information and dependencies as compared to the methods based on multi-scale feature fusion. Our model can also achieve strong performance on \textit{pedestrian}  and  \textit{cyclist} detection. The \textit{car}-BEV, \textit{pedestrian}-3D and \textit{cyclist}-3D results are presented in Table~\ref{table:add_bev_val} and Table~\ref{table:ped_cyc} for reference.

\subsection{Detection Results on the Waymo Dataset}
As for the Waymo dataset, we train our model on the training set and evaluate it on the validation set.  Likewise, the mAP is calculated with $0.7$ IoU threshold for vehicle detection. The data is split into two difficulty levels: LELVEL\_1 denotes objects containing more than 5 points , LELVEL\_2 denotes objects containing $1\sim5$ points.

\noindent\textbf{Performance Comparisons.}
In Table~\ref{table:waymo}, we compare our CT3D with state-of-the-art methods based on official released evaluation tools~\cite{sun2020scalability}. It can be seen that our method outperforms all previous methods with remarkable margins on all distance ranges of interest in both LEVEL\_1 and LEVEL\_2. CT3D achieves 76.30\% for the commonly used LEVEL\_1 3D mAP evaluation metric, surpassing previous state-of-the-art method Voxel-RCNN by 0.71\% on 3D detection, and 2.31\% on bird-view detection. 
This significant improvement also verifies the effectiveness of our CT3D approach on large-scale point cloud feature representation.
We report the results of LEVEL\_2 difficulty in Table~\ref{table:waymo}, our method outperforms Voxel-RCNN significantly by 2.45\% on 3D detection. A contributing factor is that Voxel-RCNN limits the feature interactions via dividing the RoI space into grids, while our proposed CT3D has the obvious advantage of capturing long-range interactions among sparse points.

\begin{figure}
	\begin{center}
		\includegraphics[width=0.9\linewidth]{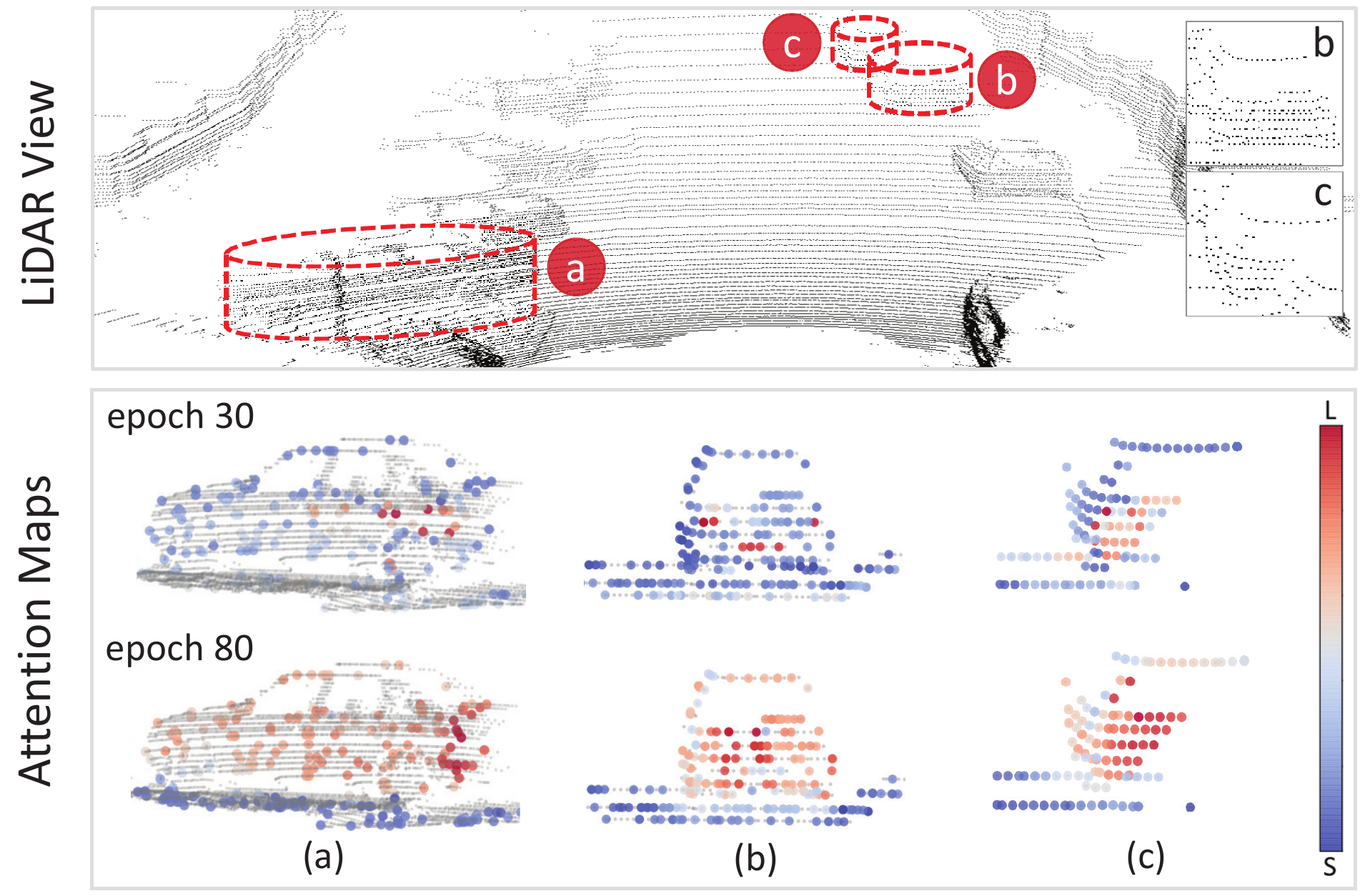}
	\end{center}
	\caption{Attention maps generated by the self-attention layer.  We visualize the weights of at most 256 sampled points within 3 RoIs (red dotted line) as the $30$-th and $80$-th epochs.}
	\label{fig:vis}
\end{figure}

\subsection{Ablation Studies}\label{ablation}
In this section, we conduct comprehensive ablation studies for the CT3D to verify the effectiveness of each individual component. We report the 3D detection AP metric with 40 recall positions on the KITTI  \textit{val} set.

\noindent\textbf{Different RPN Backbones.} In Table~\ref{table:ablation_rpn}, we validate the effects of our refinement network with 
“SECOND RPN~\cite{second},” and “PointPillar RPN~\cite{lang2019pointpillars}”, respectively. It can be seen that the detection performance boosts with +5.47\% and +4.82\% when compared to the RPN baselines. This benefits from that our two-stage framework CT3D could be integrated on the top of any RPNs to provide strong ability for proposal refinement. We also provide the amount of parameters in Table~\ref{table:ablation_rpn} for reference.

\noindent\textbf{Proposal-to-point Embedding.} We investigate the importance of the keypoints subtraction strategy by comparing it with the baseline size-orientation strategy adopted in the proposal-to-point embedding of Sec.~\ref{encoder}. The $2^{nd}$ and $3^{rd}$ rows of Table~\ref{table:ablation} show that keypoints subtraction approach significantly improves the performance in all three difficulty levels. The rationale behind this strategy is that the relative coordinates between each point and the proposal keypoints could provide more effective geometric information, forming high-quality point location embeddings.

\noindent\textbf{Self-attention Encoding.} The $1^{st}$ and $3^{rd}$ rows of Table~\ref{table:ablation} show that removing the self-attention encoding drops performance a lot, which demonstrates that the self-attention enables better feature representation for each point by aggregating the global-aware context information and dependencies. Moreover, we visualize the attention maps of the last self-attention layer of a trained model from different epoch checkpoints. As shown in Figure~\ref{fig:vis}, the points on cars get more attention in epoch 80, even in an extremely sparse case as Figure~\ref{fig:vis} (c). On the contrary, the background points get less attention with the training process. Therefore, CT3D  pays more attention to foreground points, and thus achieves considerable performance.

\noindent\textbf{Channel-wise Decoding.} 
As shown in the $3^{rd}$, $4^{th}$ and $5^{th}$ rows of Table~\ref{table:ablation}, the extended channel-wise re-weighting outperforms both the standard decoding and channel-wise re-weighting with a large margin. This benefits from the  integration of the standard decoding and the channel-wise re-weighting for both global and channel-wise local aggregation, generating more effective decoding weights.

\section{Conclusion}
In this paper, we present a two-stage 3D object detection framework CT3D with a novel channel-wise Transformer architecture. Our method first encodes the proposal information into each raw point via an efficient proposal-to-point embedding, followed by self-attention to capture the long-range interactions among points. Subsequently, we transform the encoded point features into a global proposal-aware representation by an extended channel-wise re-weighting scheme which could obtain effective decoding weights for all points. The CT3D provides a flexible and highly-effective framework which is particularly helpful for point cloud detection tasks. Experimental results on both the KITTI dataset and the large-scale Waymo dataset also verify that CT3D could achieve significant improvement over the state-of-the-art methods.

\begin{table}[t]
	\footnotesize
	\setlength\tabcolsep{5pt}
	\begin{center}
		\begin{tabular}{ccc|c||c}
			\hline
			PointPillar & SECOND  &  Two-stage & Par.& 	\multirow{2}*{Moderate AP (\%)}\\ 
			RPN & RPN &refinement & (M)& \\
			\hline
			\checkmark & &&18&79.26\\
			\checkmark &&\checkmark&28 &84.08\\
			&\checkmark & &20& 80.35\\
			&\checkmark &\checkmark&30 & \textbf{85.82}\\
			\hline
		\end{tabular}
	\end{center}
	\caption{Ablation studies for different RPNs on the KITTI \textit{val} set in terms of 3D detection AP metric with 40 recall positions. }
	\label{table:ablation_rpn}
\end{table}

\begin{table}[t]
	\footnotesize
	\setlength\tabcolsep{5pt}
	\begin{center}
		\begin{tabular}{ccccc||ccc}
			\hline
			K. S. & S. E. & S. D. & C. R. & E. C. R & Easy & Mod. & Hard\\
			\hline
			\checkmark &&\checkmark&&&90.29 & 79.20 & 74.59\\
			&\checkmark& \checkmark &&&91.92&83.41&81.79\\
			\checkmark&\checkmark  & \checkmark &&& 92.09&85.10&82.98\\
			\checkmark&\checkmark  &&\checkmark && 92.56 & 85.34 & 83.23\\
			\checkmark&\checkmark  &&& \checkmark & \textbf{92.85}&\textbf{85.82}&\textbf{83.46}\\
			\hline
		\end{tabular}
	\end{center}
	\caption{Ablation studies for proposal-to-point embedding, self-attention encoding and channel-wise decoding on the KITTI \textit{val} set. “K. S.” stands for the keypoints subtraction strategy, “S. E.” stands for the self-attention encoding, “S. D.”, “C. R.” and “E. C. R.” represent the standard decoding, channel-wise re-weighting, and our extended channel-wise re-weighting, respectively.}
	\label{table:ablation}
\end{table}

\section*{Acknowledgements}
This work was supported by Alibaba  Innovative Research (AIR) progamm and  Major Scientifc Research Project of Zhejiang Lab (No. 2019DB0ZX01).


{\small
\bibliographystyle{ieee_fullname}

}

\end{document}